\documentclass[11pt,a4paper]{article}
\usepackage[hyperref]{acl2017}
\usepackage{times}
\usepackage{latexsym}
\usepackage[english]{babel}
\usepackage{blindtext}
\usepackage{adjustbox}
\usepackage{dialogue}

\usepackage{natbib}

\usepackage{url}

\usepackage{adjustbox}
\usepackage{multirow}
\usepackage{graphicx}
\usepackage{caption}
\usepackage{blindtext}
\usepackage{xparse}
\usepackage{dialogue}
\usepackage{rotating}
\usepackage{listings}
\usepackage{mathtools}
\usepackage{pgfgantt}
\usepackage{lscape}
\usepackage{subcaption}
\usepackage{cleveref}
\usepackage{url}
\usepackage{boxedminipage}

\aclfinalcopy 

\definecolor{darkgreen}{rgb}{0.0, 0.5, 0.0}

\makeatletter
\def\ODhl#1{\bgroup \markoverwith{\lower3.5\p@\hbox{\sixly \textcolor{darkgreen}{\char58}}}\ULon{#1}}
\font\sixly=lasyb10 scaled 652 
\makeatother
\def\ODdel#1{\bgroup\markoverwith{\textcolor{darkgreen}{\rule[0.5ex]{2pt}{1pt}}}\ULon{#1}}

\title{Using Emotion and Dialogue-related Features\\to Predict Human Ratings of Conversational Robots}
\title{Using Multimodal Features to Evaluate \\Spoken Human-Robot Dialogue Interaction}
\title{Sympathy Begins with a Smile, Intelligence Begins with a Word:\\ Use of Multimodal Features in Spoken Human-Robot Interaction}

\author{Jekaterina Novikova, Christian Dondrup, Ioannis Papaioannou \and Oliver Lemon\\
Interaction Lab \\
Heriot-Watt University \\
Edinburgh, EH14 4AS, UK \\
  {\tt \{j.novikova, c.dondrup, i.papaioannou, o.lemon\}@hw.ac.uk}}

\date{}

\begin{document}

\maketitle
\begin{abstract}
Recognition of social signals,   from human facial expressions or prosody of   speech, is a popular research topic in human-robot interaction studies. There is also a long line of research in the spoken dialogue community that investigates user satisfaction in relation to dialogue characteristics. However, very little research  relates a combination of multimodal social signals and language features detected during spoken face-to-face human-robot interaction to the resulting user perception of a robot. 
In this paper we show how different emotional facial expressions of human users, in combination with prosodic characteristics of human speech and features  of human-robot dialogue, correlate with  users' impressions of the robot after a conversation. We find that {\it happiness} in the user's recognised facial expression strongly correlates with likeability of a robot, while {\it dialogue-related} features (such as number of human turns or number of sentences per robot utterance) correlate with perceiving a robot as   intelligent.
In addition, we show that    facial expression, emotional features, and prosody are better  predictors of human ratings related to perceived robot likeability and anthropomorphism, while linguistic and non-linguistic features more often predict perceived robot intelligence and interpretability. As such, these characteristics may in future be used as an online reward signal for in-situ Reinforcement Learning-based adaptive human-robot dialogue systems.

\end{abstract}

\begin{figure}[t]
\centering
    \begin{subfigure}[b]{0.26\textwidth}
        \includegraphics[width=\textwidth]{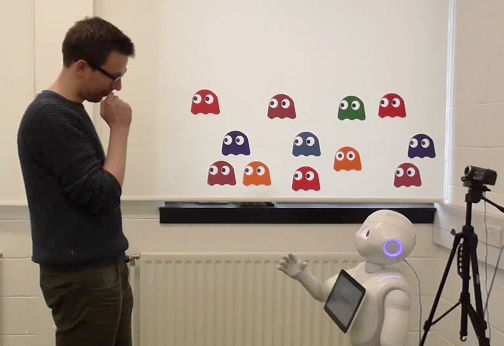}
        \label{fig:experiment_set-up_live}
    \end{subfigure}
    \hfill
    \begin{subfigure}[b]{0.15\textwidth}
        \includegraphics[width=\textwidth]{exp_setup.png}
        \label{fig:experiment_set-up_diagram}
    \end{subfigure}    
\caption{Left: a live view of experimental setup showing a participant interacting with Pepper. Right: a diagram of experimental setup showing the participant (green) and the robot (white) positioned face to face. The scene was recorded by cameras (triangles C) from the robot's perspective focusing on the face of the participant and from the side, showing the whole scene. The experimenter (red) was seated behind a divider.}
\label{fig:experiment_set-up}
\vspace{-.5cm}
\end{figure}

\section{Introduction}

Social signals, such as emotional expressions, play an important role in human-human interaction, thus they are increasingly recognised as an important factor to be considered both in human-robot interaction research \cite{cid2013real, novikova2015emotionally, devillers2015inference} and in the area of spoken dialogue systems \cite{herm2008calls, meena2015automatic}. 

Recognition of human social signals has become a popular topic in Human-Robot Interaction (HRI) in recent years. Social signals are recognized well from human facial expressions or prosodic features of speech \cite{ekman2004emotional,zeng2009survey}, and have become the most popular methods for recognising human affective signals   in human-robot interaction \cite{razuri2015speech,devillers2015inference,cid2013real}. 

In human-robot interaction, recognized human emotions are mostly used for mimicking human behaviour and enhancing the empathy towards a robot both in children \cite{tielman2014adaptive} and in adult users \cite{tapus2007emulating}. 

In the area of spoken dialogue systems, signals recognised from linguistic cues and prosody have been used to detect problematic dialogues \cite{herm2008calls} and to assess dialogue quality as a whole \cite{schmitt2015interaction}. This type of dialogue-related signals has also been used to automatically detect miscommunication \cite{meena2015automatic}, or to predict the user satisfaction~\cite{schmitt2011modeling}.

However, there is very little research combining the areas of detecting multi-modal signals during spoken HRI and evaluation of human-robot conversation, and using them to create an adaptive social dialogue. 

In this paper, we make a first step towards building a multi-modally-rich, conversational, and human-like robotic agent, potentially able to react to the changes in human behaviour during face-to-face dialogue and able to adjust the dialogue strategy in order to improve an interlocutor's impression. We present a setup that targets the development of a dialogue system to explore verbal and non-verbal conversational cues in a face-to-face situated dialogue with a social robot. We show that different emotional facial expressions of a human interlocutor, in combination with prosodic characteristics of human speech and features of human-robot dialogue, correlate strongly with users' perceptions of a robot after a conversation. Based on these features, we developed a model capable of predicting potential human ratings of a robot and discuss its implications for future work in developing adaptive human-robot dialogue systems.

\section{Experiment Setup and Evaluation} 
\label{sec:eval}

The human-robot dialogue system was evaluated via a user study in which human subjects interacted with a Pepper robot\footnote{\url{http://doc.aldebaran.com/2-5/home_pepper.html}} acting autonomously using the system described in \cite{Papaioannou:2017:CCT:3029798.3034820, Papaioannou2017b}. The dialogue system used, combines task-based with chat-based dialogue features, deciding the most appropriate action on each consequent turns, using a pre-trained Reinforcement Learning (RL) policy. The robot decides among a pool of possible actions $a_t \in A$ where $A =$ [\textit{PerformTask, Greet, Goodbye, Chat, GiveDirections, Wait, RequestTask, RequestShop}]. If a task is recognised in the user utterance (e.g. "where can I find discounts"), a response is synthesized using database lookup and predefined utterances (like the example shown in Table \ref{tab:ex-shorten}). If no task was recognised, then the user request is being forwarded to a Chatbot, written in AIML and based on the chatbot \textit{Rosie}\footnote{\url{http://github.com/pandorabots/rosie}}, where a chat-style response is formulated based on AIML template/ pattern matching.

All interactions were in English. The physical setup of the experiment can be seen in Figure~\ref{fig:experiment_set-up}.

\subsection{Experimental Scenario}

The task and the setup chosen in the study were considered as first steps towards understanding how a humanoid social robot should behave in the context of a shopping mall  while also providing useful information to the mall's visitors. 
To this end, participants were asked to imagine that they were entering a shopping mall they had never been to before where the robot was installed in the entry area interacting with visitors one at a time. Participants were asked to complete as many as possible of the following five tasks: 

\begin{itemize}
\item Get information from the robot on where to get a coffee.

\item Get information from the robot on where to buy clothes. 

\item Get the directions to the clothing shop of their choice.

\item Find out if there are any current sales or discounts in the shopping mall and try to get a voucher from the robot.

\item Make a selfie with the robot.
\end{itemize}

Instructions were given to use natural language spontaneously while interacting with the robot.

\subsection{Participants and Experimental Design}

41 people (13 females, 28 males) participated in our study, ranging in age from 18 to 38 (M=24.46, SD=4.72). The majority of them were students (93\% students and 7\% staff) that had no or little previous experience with robots (56\% with little or no experience, 39\% with some experience, and 5\% with a lot of experience). 

Participants were initially given a briefing script describing the goal of the task and providing hints on how to better communicate with the robot, e.g.\ ``wait for your turn to speak'' and ``please keep in mind that the robot only listens to you while its eyes are blinking blue''\footnote{Pepper's default way of communicating that it is listening.}. We reassured our participants that we were testing the robot, not them, and controlled environment-introduced biases by avoiding non-task-related distractions during the experiment. During experimental sessions, participants stood in front of the robot and the experimenter was hidden in another corner of the room but available in case the participant would need any help (see Figure~\ref{fig:experiment_set-up}).

At the end of the experiment participants were debriefed and received a \textsterling10 gift voucher. The duration of each session did not exceed thirty minutes.

\subsection{Measured Variables}

We collected a range of objective measures from the log files, video and audio recordings of the interactions, and transcripts of dialogues. From the audio recordings, we collected a set of different prosodic and dialogue-related features. From the video recordings, we collected the data on emotional intensities detected based on human facial expressions. From the dialogue transcripts, we collected a set of linguistic features, such as lexical diversity, length of utterance etc.

In addition, we considered a range of subjective measures for a qualitative evaluation. For that, after each interaction session participants were asked to fill in a questionnaire to assess their perception of the robot. 

\begin{figure}
    \centering
    \begin{subfigure}[b]{0.11\textwidth}
        \includegraphics[width=\textwidth]{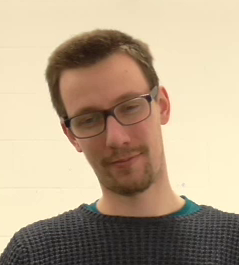}
    \end{subfigure}
    \begin{subfigure}[b]{0.11\textwidth}
        \includegraphics[width=\textwidth]{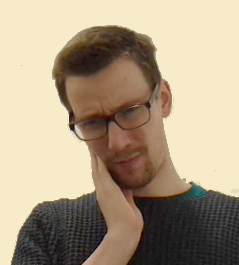}
    \end{subfigure}
    \begin{subfigure}[b]{0.11\textwidth}
        \includegraphics[width=\textwidth]{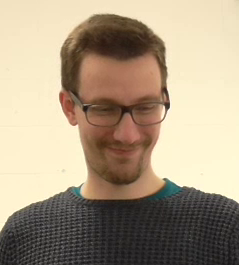}
    \end{subfigure}
    \begin{subfigure}[b]{0.11\textwidth}
        \includegraphics[width=\textwidth]{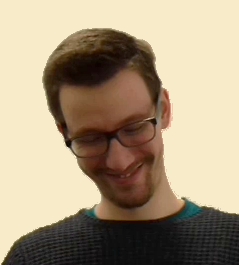}
    \end{subfigure}
    \caption{Screenshots of the recorded video, showing different facial expressions detected during a dialogue with the robot.}
    \label{fig:face-expr}
\end{figure}

\textbf{Emotions} were detected and recognised using the Microsoft Emotion API for Video\footnote{https://www.microsoft.com/cognitive-services/en-us/emotion-api}. This API takes video frames as an input (see Figure~\ref{fig:face-expr}), and returns the confidence across a set of emotions for the group of faces in the image over a period of time. The emotions detected are {\it happiness, sadness, surprise, anger, fear, contempt, disgust, or neutral}. Happiness, surprise, and sadness were selected for analysis in this work, 
because they had the highest average or maximum values across all recorded videos. 

\textbf{Prosodic Features} used in this work are the following: average fundamental frequency of speech F0, maximum F0, and difference between maximum and minimum F0 values.

\textbf{Non-linguistic Dialogue Features} used in this work contain speech duration (in sec), number of turns, number of completed tasks, number of self-repetitions and a ratio of tasks per turn.

\textbf{Linguistic Dialogue Features} used in this work consist of utterance length (in characters), a ratio of words per utterance and unique words per utterance, number of sentences within an utterance, lexical diversity, a ratio of words per sentence and a ratio of unique words per sentence.

\textbf{Perception of Robot} was assessed using responses on the questionnaire filled by participants at the end of each interaction session. The questionnaire was based on a combination of the User Experience Questionnaire UEQ \cite{laugwitz2008construction} and the Godspeed Questionnaire \cite{bartneck2009measurement}. It consisted of 21 pairs of contrasting characteristics that may apply to the robot, and are grouped into four groups of {\it Anthropomorphism, Likeability, Perceived Intelligence, and User Expectations}. The Anthropomorphism group consists of the following pairs of characteristics: {\it fake -- natural, machinelike -- humanlike, unconscious -- conscious, artificial -- lifelike}. Likeability consists of: {\it unfriendly -- friendly, unkind -- kind, unpleasant -- pleasant, awful -- nice, annoying -- enjoyable, disliked -- liked}. The group of Perceived Intelligence consists of: {\it incompetent -- competent, ignorant -- knowledgeable, irresponsive -- responsive, unintelligent -- intelligent, foolish -- sensible}. The Interpretability group consists of: {\it does not meet expectations -- meets expectations, obstructive -- supportive, unpredictable -- predictable, confusing -- clear, complicated -- easy, not understandable -- understandable}. Users were asked to evaluate perception of a robot on a 5-point Likert scale, where the minimum value was 1 and the maximum was 5.

The validity of the used questionnaire was tested by measuring its internal consistency with Cronbach's $\alpha$, which was equal to 0.93 (high consistency). Based on the high value of the Cronbach's $\alpha$, we assume that that our participants in the given context interpreted the robot characteristics, provided in the questionnaire, in an expected way.

\begin{figure}[tp]
\centering
\includegraphics[width=.45\textwidth]{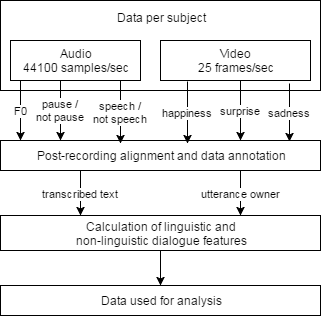}
\caption{A Flow chart showing the process of synchronising different streams of data, and collecting corresponding parts of data for analysis.}
\label{fig:data_process}
\vspace{-.5cm}
\end{figure}

\section{Multimodal Data Collection and Analysis}

Data collected during the experiment required additional processing, alignment and annotation, as shown in Figure~\ref{fig:data_process}. Prosodic features of F0, and dialogue-related features showing presence and absence of pauses and presence/absence of speech were collected from audio recordings with a rate of 44100 samples per second. Values of emotional intensities were collected from video recordings with a rate of 25 frames per seconds. All the data was aligned after recording, using average values of prosodic features  per frame. Afterwards, data was annotated in ELAN\footnote{https://tla.mpi.nl/tools/tla-tools/elan} detecting associations between an utterance and its owners. Finally, the dialogue texts were transcribed and linguistic features were calculated using R packages \textit{stringr, stringi, tidytext}, and \textit{qdap}.

\begin{table}[tp]
\centering
\begin{adjustbox}{max width=0.49\textwidth}
\begin{tabular}{llll}
\begin{tabular}[c]{@{}l@{}}\textbf{Group of}\\ \textbf{features}\end{tabular} & \textbf{Feature} & \textbf{Human} & \textbf{Robot} \\
 \hline \hline
\multirow{3}{*}{\begin{tabular}[c]{@{}l@{}}Emotional\\ features\end{tabular}} & Happiness & 0.40 & NA \\
 & Surprise & 0.01 & NA \\
 & Sadness & 0.01 & NA \\
 \hline
\multirow{3}{*}{\begin{tabular}[c]{@{}l@{}}Prosodic\\ features\end{tabular}} & F0, avg & 173.87 & NA \\
 & F0, max & 398.63 & NA \\
 & F0, diff & 338.49 & NA \\
\hline
\multirow{7}{*}{\begin{tabular}[c]{@{}l@{}}Linguistic\\ dialogue-\\ related \\ features\end{tabular}} & Utterance length & 21.34** & \textbf{26.78}** \\
 & Words / Utterance & 5.74** & \textbf{7.16}** \\
 & Unique words / Utterance & 5.49* & \textbf{6.70}* \\
 & Lexical Diversity & \textbf{0.97} & 0.95 \\
 & No of sentences & 1.13 & \textbf{1.24} \\
 & Words / Sentence & 5.19 & \textbf{5.99}* \\
 & Unique words / Sentence & 5.03 & \textbf{5.63} \\
\hline
\multirow{5}{*}{\begin{tabular}[c]{@{}l@{}}Non-linguistic\\ dialogue-\\ related\\ features\end{tabular}} & Speech duration, sec & 36.54 & \textbf{48.92}** \\
 & No of turns & \textbf{32.81} & 29.38 \\
 & No of completed tasks & 4.00 & NA \\
 & No of self-repetitions & 3.13 & \textbf{4.19} \\
 & Tasks / Turn & \textbf{0.16} & 0.15
\end{tabular}
\end{adjustbox}
\caption{Descriptive statistics of emotional, prosodic, non-linguistic and linguistic dialogue features for human and robot actors. Here, bold indicates a higher value, ** denotes p $<$ 0.01, * denotes p $<$ 0.05}
\label{tab:summary}
\end{table}

A summary of collected data is provided in Table~\ref{tab:summary}. 
Specifically, the summary results show that the F0 value of human speech changes a lot during the conversation, with a maximum value being more than twice as large as an average value. Average emotional intensities of surprise and sadness, on the other hand, do not differ much $(\pm0.001)$, and the maximum values of all the emotional intensities are usually close to 1. 

Results of non-linguistic dialogue-related features show that the robot on average speaks significantly longer than humans during a dialogue. Humans tend to have a higher number of turns, although they less frequently repeat themselves. These differences, however, are not significant. 

Results of linguistic features reveal more significant differences between robot and human language. For example, the results show that humans on average speak in significantly shorter utterances compared to a robot, both in terms of a number of characters and a number of words per utterance. The robot uses more sentences per utterance on average, although this difference is not significant. The lexical diversity, which was calculated as a ratio of unique words and a total number of words in an utterance, shows a slightly higher value in human language rather than robot's. 

Values of linguistic features differ significantly between human and robot language, which leads us to investigate in more details the textual dialogue data in terms of lexical variety and syntactic complexity.

\section{Linguistic Analysis of Dialogues}

Following \citet{gardent_creating_2017}, we analyse the dialogue textual data in terms of length of utterances, lexical richness, and syntactic variation. The results are summarised in Tables~\ref{tab:summary} and~\ref{tab:ling-data} and grouped by a speaker, i.e.\ robot and human.

\begin{table}[tp]
\centering
\begin{adjustbox}{max width=0.49\textwidth}
\begin{tabular}{l|ll|l}
Speaker & LS & MSTTR & \begin{tabular}[c]{@{}l@{}}D-level \\ complexity\end{tabular} \\
 \hline \hline
robot & 0.44 & \textbf{0.61} & \textbf{1.71*} \\
human &  \textbf{0.47*} & 0.59 & 1.68
\end{tabular}
\end{adjustbox}
\caption{Results of linguistic dialogue analysis. * denotes p $<$ 0.05.}
\label{tab:ling-data} 
\end{table}

\subsection{Length of utterances} 

Results presented in Table~\ref{tab:summary} show that robot utterances are significantly longer than those of their human interlocutors, both in terms of words per utterance and sentences per utterance. This may be partly explained by the fact that a turn-taking process was not very natural and thus was not always successful during the dialogue. It usually took some time for people to learn how to communicate with Pepper properly and to start speaking to the robot only when it was listening. As a result, from time to time people were interrupted by the robot, while they never tried to interrupt the robot themselves. Shorter average length of human utterances is also caused by the way people tend to deal with disfluencies of a dialogue, e.g.\ rephrasing and shortening their previous utterance in order to emphasise the most important keywords (see an example in Table~\ref{tab:ex-shorten}). The robot utterances, on the other hand, were not shortened or changed in any other way in the case of dialogue disfluencies. 

\begin{table}[ht]
\begin{dialogue}
\speak{Human (H)} By the way, I'm a student so I don't have a lot of money. So, is it possible to have some shop with sales or discounts? [30 words]
\speak{Robot (R)} Who, specifically, does? [dialogue disfluency]
\speak{H} To have some discounts somewhere. [5 words]
\speak{R} There are 2 shops that have sales nearby. These are Tesco,  and Phone Heaven.
\speak{H} Thank you very much.
\end{dialogue}
\caption{An example of shortening as a result of dialogue disfluency.}
\label{tab:ex-shorten}
\end{table}

\subsection{Lexical Richness}

We used the Lexical Complexity Analyser \cite{lu2009automatic} 
to measure various dimensions of lexical richness, such as lexical sophistication, lexical diversity and mean segmental type-token ratio. 
We complement the traditional measure of lexical diversity  {type-token ratio} (TTR) with the more robust measure of {mean segmental type-token ratio} (MSTTR) \cite{lu2012relationship}, which divides all the dialogues into successive segments of a given length and then calculates the average TTR of all segments. The higher the value of MSTTR, the more diverse is the measured text. We also measure {\em lexical sophistication} (LS), also known as lexical rareness, which is calculated as the proportion of lexical word types not on the list of 2,000 most frequent words generated from the British National Corpus. In addition, we measure \textit{lexical diversity} (LD) as a ratio of unique and total words per utterance. 

\begin{figure}[t]
\includegraphics[width=0.45\textwidth]{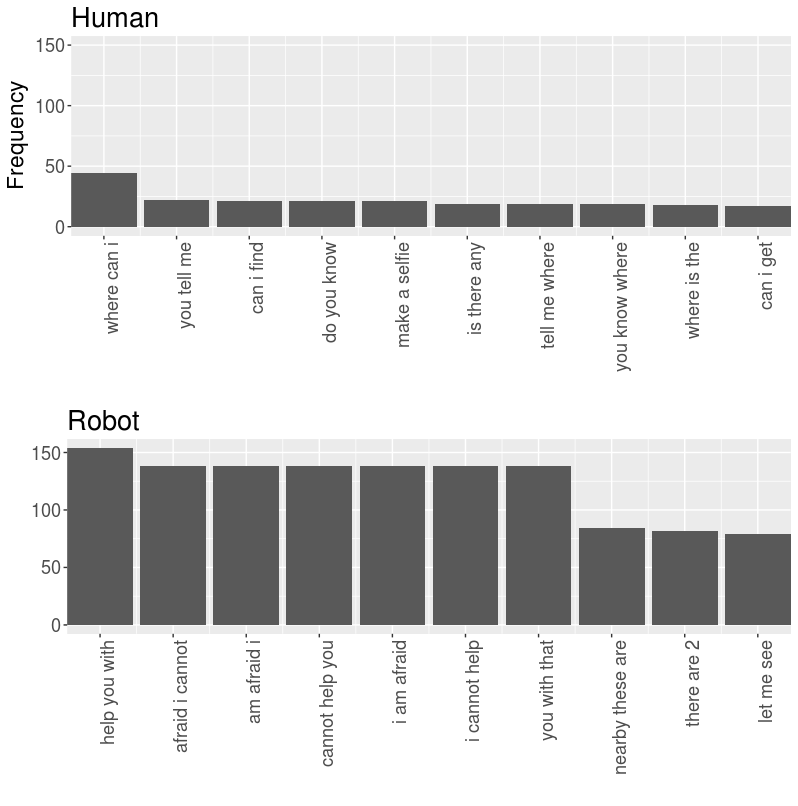}
\caption{Distribution of the top-10 most frequent trigrams in human and robot language.}
\label{fig:trigrams}
\end{figure}

The results presented in Table~\ref{tab:ling-data} show that human utterances, although being significantly shorter, are significantly richer than those of the robot, both in terms of lexical diversity and lexical sophistication. MSTTR values do not differ significantly between human and robot utterances. This leads us to investigate the distribution of frequencies of bigrams and trigrams in human and robot utterances during dialogues. 

The majority of both robot (61\%) and human (62\%) bigrams are only used once in all the dialogues. However, the mean frequency of bigrams that were used more than once during dialogues is significantly (p $<$ 0.001) higher in robot utterances (Mean = 15.8, SD = 31.4) compared to human utterances (Mean = 6.1, SD = 8.4). This means that the robot tends to use the same combinations of words repeatedly, while people do vary their language more. The majority of trigrams
 is also used just once by both people and the robot,
 although the proportion is quite different: 75\% of human trigrams and only 65\% of robot trigrams are used once in the dialogues. Those trigrams that are used more than once, have an average frequency of 15.8 (SD = 30.9) for robot, and only 4.4 (SD = 4.6) for human utterances. 

The results of bigrams and trigrams analysis support the conclusion that human language in human-robot conversations is more rich, varied, and diverse than that of the robot. Figure~\ref{fig:trigrams} shows that poor lexical variation of a robot language is influenced a lot by the fact that the robot often uses the phrase ``I am afraid I cannot help you with that", which may be said when the speech recognition confidence does not reach an adequate threshold, or when no known keywords are detected in human utterances. 
As Figure~\ref{fig:trigrams} shows, 7 out of 10 most frequent trigrams in a robot language are variations of that specific phrase.

 \subsection{Syntactic Variation and Discourse Phenomena}
 
We used the D-Level Analyser \cite{lu2009automatic}
to evaluate syntactic variation and complexity of human references using the revised D-Level Scale~\cite{lu2014computational}. 
The scale has eight levels of syntactic complexity, where levels 0 and 1 include simple or incomplete sentences and higher levels include sentences with more complex structures.

\begin{figure}[t]
\centering
\includegraphics[width=.45\textwidth]{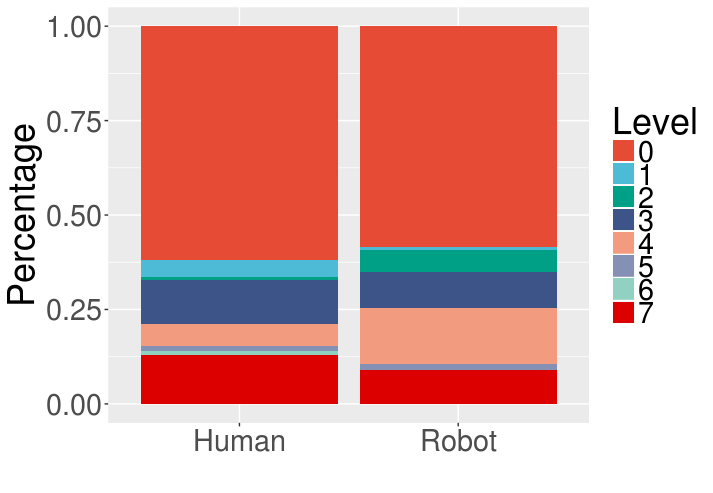}
\vspace{-.5cm}
\caption{D-level sentence distribution of human and robot language.}
\label{fig:dlevel}
\vspace{-.2cm}
\end{figure}

\begin{table*}[tp]
\centering
\begin{adjustbox}{max width=0.98\textwidth}
\begin{tabular}{ll|ll}
\textbf{} & \textbf{} & \multicolumn{2}{c}{\textbf{Significant correlation with human ratings (Spearman)}} \\
\textbf{\textit{Group of features}} & \textbf{\textit{Features}} & \textbf{\textit{\begin{tabular}[c]{@{}l@{}}Feature calculated for \\ Human\end{tabular}}} & \textbf{\textit{\begin{tabular}[c]{@{}l@{}}Feature calculated for \\ Robot\end{tabular}}} \\
\hline \hline
\multirow{3}{*}{\begin{tabular}[c]{@{}l@{}}Emotional \\ features\end{tabular}} & Happiness & Friendly (0.72), Nice (0.67) & -- \\
 & Surprise & NA & -- \\
 & Sadness & Sensible (0.71) & -- \\
 \hline
\multirow{3}{*}{\begin{tabular}[c]{@{}l@{}}Prosodic \\ features\end{tabular}} & F0, avg & Knowledgeable (0.57) & -- \\
 & F0, max & NA & -- \\
 & F0, diff & NA & -- \\
 \hline 
\multirow{5}{*}{\begin{tabular}[c]{@{}l@{}}Non-linguistic\\ dialogue-\\ related\\ features\end{tabular}} & Speech duration, sec & NA & NA \\
 & No of turns & Intelligent (0.64), Knowledgeable (0.54) & Disliked (0.54) \\
 & No of self-repetitions & \begin{tabular}[c]{@{}l@{}}Clear (0.65), Easy (0.52), \\ Understandable (0.50)\end{tabular} & \begin{tabular}[c]{@{}l@{}}Awful (0.58), \\ Does not meet expectations (0.52), \\ Annoying (0.49)\end{tabular} \\
 & No of completed tasks & Clear (0.63), Meets expectations (0.58) & --  \\
 & Tasks / Turn & Ignorant (0.58), Unintelligent (0.52) & \begin{tabular}[c]{@{}l@{}}Humanlike (0.63), Unintelligent (0.55), \\ Ignorant (0.53)\end{tabular} \\
 \hline 
\multirow{7}{*}{\begin{tabular}[c]{@{}l@{}}Linguistic\\ dialogue-\\ related\\ features\end{tabular}} & Utterance length & Responsible (0.54) & Annoying (0.62), Obstructive (0.50) \\
 & Words / Utterance & NA & Annoying (0.62), Obstructive (0.57) \\
 & Unique words / Utterance & NA & Annoying (0.62), Obstructive (0.57) \\
 & Lexical Diversity & \begin{tabular}[c]{@{}l@{}}Conscious (0.61), Humanlike (0.58), \\ Natural (0.56)\end{tabular} & Supportive (0.54) \\
 & No of sentences & Responsible (0.50) & Easy (0.50), Ignorant (0.57) \\
 & Words / Sentence & NA & Competent (0.52) \\
 & Unique words / Sentence & Confusing (0.54) & Competent (0.56)
\end{tabular}
\end{adjustbox}
\caption{Summary of correlation between a mean value of feature, calculated during a dialogue, and human ratings of the robot. Only significant correlations are included. ``NA" means no significant correlation was observed, ``--" means that correlation was not calculated.}
\label{tab:corr}
\end{table*}

Figure~\ref{fig:dlevel} shows a similar syntactic variation in human and robot language, although there are slight differences, e.g.\ people tend to use a higher percentage of both the simplest and the most complicated sentences. In general, the majority of all the sentences, used both by humans and by a robot, are simple sentences. This is because the topic of a human-robot conversation is quite simple and does not require a lot of complicated syntactic structures. 

The results of initial linguistic analysis, together with results of analysis of multimodal signals, suggest that linguistic, as well as other multimodal features, may be important in predicting human perception of a robot. However, average scores can be misleading, as they only provide a system-level overview but do not measure the strength of association with human ratings. This led us to inspect the correlation between the ratings of the robot and all the multimodal features of a dialogue. 

\section{Correlation between Robot Ratings and Multimodal Features of Human-Robot Dialogue}

A summary of correlation results is presented in Table~\ref{tab:corr}. The results reveal that different groups of features correlate with different groups of human ratings. For example, emotional features, such as intensity of happiness, correlate strongly with perceived anthropomorphism of a robot, so that a person who more strongly expresses happiness during a dialogue with the robot probably perceives it as friendlier and nicer. Human ratings of perceived robot anthropomorphism also correlate with a lexical diversity of human language: people tend to use more diverse language when speaking to a robot  that they perceive as conscious, natural, and humanlike (see an example in Table~\ref{tab:ex-anthropomorphismH}).  

\begin{table*}[ht]
\begin{minipage}{.49\textwidth} %
\begin{dialogue}
\speak{H} Where is Costa? Is it far? [LD = 0.83]
\speak{R} A place we call ``reality". [disfluency]
\speak{H} Sorry I didn't understand. [LD = 1.0]
\speak{R} There is no need to apologize. Perhaps I can explain it better. Which part are you confused about?
\end{dialogue} 
\end{minipage} 
\hfill
\begin{minipage}{.49\textwidth} %
\begin{dialogue}
\speak{H} Bye bye [LD = 0.50]
\speak{R} My birthday is September 9th. [disfluency]
\speak{H} Bye bye [LD = 0.50]
\speak{R} Would you like to do a search for it? [disfluency]
\speak{H} No. [LD = 1.0]
\end{dialogue} 
\end{minipage}
\caption{An example of a dialogues that resulted in: Left - high human ratings for anthropomorphism, Right - low human ratings for anthropomorphism. Both dialogues present situations when human intent was not recognised by the robot. LD denotes lexical diversity.}
\label{tab:ex-anthropomorphismH}
\end{table*}

Average F0 value of human speech correlates strongly with perceived intelligence of the robot, specifically with a robot being more knowledgeable. The ratio of tasks per turn is, unsurprisingly, strongly correlated with perceived robot intelligence. The more dialogue turns  people need to complete the same number of tasks, the more they perceive the robot as ignorant and unintelligent. Features of robot language also correlate with how it is perceived: the more words (including unique ones) per sentence it generates, the more competent it appears to humans.

It is interesting to notice that some linguistic and non-linguistic dialogue features correlate with different human ratings depending on whether the features are calculated for human or robot language. For example, a higher number of human turns during a dialogue correlates strongly with a robot being perceived as intelligent and knowledgeable, while a higher number of robot turns correlates with it being disliked. Longer human sentences show that a robot is perceived as more responsive, while longer robot sentences correlate with a robot being annoying and obstructive.

The results show that some features, observable during a human-robot dialogue, correlate strongly and significantly with different groups of human ratings. However, it is not obvious if a strong correlation also means that there is a causal relationship between human language or multimodal behavioural features and ratings of the robot. 
This leads us to inspect whether the previously discussed features may be used for predicting potential ratings. 

\section{Predicting Perception of Robots in Human-Robot Dialogue}

\begin{table*}[tp]
\centering
\begin{adjustbox}{max width=0.98\textwidth}
\begin{tabular}{l|llllll|l}
\textit{\textbf{Group of rating}} & \textit{\textbf{\begin{tabular}[c]{@{}l@{}}Emotions\\ only\end{tabular}}} & \textit{\textbf{\begin{tabular}[c]{@{}l@{}}Prosody\\ only\end{tabular}}} & \textit{\textbf{\begin{tabular}[c]{@{}l@{}}Non-linguistic\\ only\end{tabular}}} & \textit{\textbf{\begin{tabular}[c]{@{}l@{}}Linguistic\\ only\end{tabular}}} & \textit{\textbf{\begin{tabular}[c]{@{}l@{}}All\\ combined\end{tabular}}} & \textit{\textbf{Baseline}} & \textit{\textbf{\begin{tabular}[c]{@{}l@{}}Average\\ rating\end{tabular}}} \\
\hline \hline
Likeability & \textbf{0.85} & 1.03 & 0.96 & 0.87 & 1.00 & 1.41 & 4.01 \\
\hline
\begin{tabular}[c]{@{}l@{}}Perceived\\ Intelligence\end{tabular} & 0.73 & 0.94 & \textbf{0.69} & 0.83 & 0.89 & 0.83 & 3.44 \\
\hline
Interpretability & 0.71 & 0.87 & \textbf{0.56} & \textbf{0.68} & 0.81 & 0.92 & 3.56
\end{tabular}
\end{adjustbox}
\caption{Performance of prediction, calculated using root-mean-square error (RMSE). The results are averaged over all the ratings that belong to the group and outperform the baseline. Bold denotes the smallest average error and means the best predicted result of the model.}
\label{tab:predict}
\end{table*}

In order to develop a model that predicts potential human ratings on robot likeability and perceived intelligence, we use the previously discussed prosodic features, dialogue-related characteristics, and detected emotional intensities, as predictive features of the model. 
For the prediction itself, we use ensemble learning (Random Forest, RF) \cite{breiman2001random} which is a state-of-the-art algorithm that can be applied in a dynamic dialogue situation and is able to combine the respective strengths of different informative features into a single model.

{\bf Setup:} We use a 70/30\% split for training and testing and 10-fold cross-validation on the training data to tune the optimal number of predictors selected for growing trees. 100 trees were grown with 2 variables randomly sampled as candidates at each split. We investigate five different models used as predictors: 1) emotional intensities, 2) prosodic features, 3) non-linguistic dialogue features, 4) linguistic dialogue features, and 5) all the features combined.

\textbf{Results:} The results in Table~\ref{tab:predict} show that different groups of features are better predictors of different groups of ratings. For example, combining dialogue-related features only (either linguistic or non-linguistic) as predictors,  produces the lowest root-mean-square error (RMSE) for many ratings out of the perceived intelligence and intelligibility groups. This means that a combination of dialogue-related features is producing the best prediction of such aspects of perceived robot intelligence as e.g.\ responsiveness, intelligence, or predictability.

Emotional features are shown to be the best predictors of some aspects of robot likeability and perceived anthropomorphism. For example, the ratings for \textit{unconscious-conscious, unfriendly-friendly} or \textit{awful-nice} are best predicted using emotional features only. Other aspects of robot anthropomorphism and likeability, such as \textit{machinelike-humanlike} or \textit{disliked-liked}, are best predicted by using only prosodic features of human speech as predictors.   
Combining emotional, prosodic and dialogue-related features rarely improves the results of rating predictions. In some cases, e.g.\ predicting the ratings for \textit{unresponsive-responsive}, a combination of all the features produces the same results as dialogue-related features alone. In one case a combination of all features does improve prediction results, this is the rating showing if the robot meets human expectations or not. This is probably because human expectations consist of different aspects themselves: some expect the robot to be anthropomorphic and likeable, other prefer it to be intelligent and easily interpretable. 

\section{Discussion and Conclusions}

In this paper, we show how dialogue features correlate with the user's perception of a robot (e.g.\ strong correlation between higher number of human turns and higher robot's perceived intellect, or between higher number of sentences per robot utterance and robot's perceived ignorance), as well as correlations between emotional features and robot likeability.

Using the findings described in this paper, a predictive model could be implemented using emotional intensities (\textit{happiness, sadness, and surprise}) in order to better predict the user's perception of the robot. This model can provide valuable information on how to design  more engaging dialogues between robots and humans. The combination of these emotional features, along with the dialogue-related features (both linguistic and non-linguistic) and the F0 value can also provide better feedback in cases where, for instance, a smile can create ambiguity of the perceived user's emotional display~\cite{DBLP:journals/corr/abs-1203-0699}.

In future work, these emotional features coming from real-time facial expression recognition  could be used as an online estimator of how well or badly a dialogue is progressing, which would be an important component of a reward signal for Reinforcement Learning approaches to HRI.



\newpage
\bibliographystyle{acl_natbib}

\bibliography{robonlp}

\end{document}